\documentclass[letterpaper, 10 pt, conference]{ieeeconf}  % Comment this line out
\IEEEoverridecommandlockouts                              % This command is only
\usepackage{graphicx}
\usepackage{svg}
\usepackage{amsmath} % assumes amsmath package installed
\usepackage{amssymb}  % assumes amsmath package installed

\usepackage{microtype}
\usepackage{hyperref}
\usepackage{textcomp}
\usepackage[ruled,vlined,linesnumbered]{algorithm2e}
\SetNlSty{texttt}{}{.}
\usepackage{tabulary}
\usepackage{multirow}% http://ctan.org/pkg/multirow
\usepackage{bm}
\usepackage{caption}
\DeclareMathOperator{\trace}{tr}

\title{\LARGE \bf
Learning to Live Life on the Edge: \\ Online Learning for Data-Efficient Tactile Contour Following
}

\author{Elizabeth A.\ Stone\textsuperscript{1}, Nathan F.\ Lepora\textsuperscript{2} and David A.W.\ Barton\textsuperscript{3}% <-this % stops a space
\thanks{\textsuperscript{1}ES is a PhD student at the EPSRC Centre for Doctoral Training in Future Autonomous and Robotic Systems (FARSCOPE) at the Bristol Robotics Laboratory. \texttt{\small lizzie.stone@brl.ac.uk}}%
\thanks{\textsuperscript{2}NL  is with the Dept.\ of Engineering Maths and Bristol Robotics Laboratory, University of Bristol, UK, and is supported by the Leverhulme Trust (RL-2016-39). \texttt{\small n.lepora@bristol.ac.uk}}
\thanks{\textsuperscript{3}DB is with the Dept.\ of Engineering Maths, University of Bristol, UK.
        \texttt{\small david.barton@bristol.ac.uk}}%
}

\begin{document}

\maketitle
\thispagestyle{empty}
\pagestyle{empty}

%%%%%%%%%%%%%%%%%%%%%%%%%%%%%%%%%%%%%%%%%%%%%%%%%%%%%%%%%%%%%%%%%%%%%%%%%%%%%%%%
\begin{abstract}
  Tactile sensing has been used for a variety of robotic exploration and manipulation tasks but a common constraint is a requirement for a large amount of training data. This paper addresses the issue of data-efficiency by proposing a novel method for online learning based on a Gaussian Process Latent Variable Model (GP-LVM), whereby the robot learns from tactile data whilst performing a contour following task thus enabling generalisation to a wide variety of tactile stimuli. The results show that contour following is successful with comparatively little data and is robust to novel stimuli. This work highlights that even with a simple learning architecture there  are  significant  advantages  to  be  gained in efficient and robust task performance by  using  latent variable  models  and  online learning for tactile sensing  tasks. This paves the way for a new generation of robust, fast, and data-efficient tactile systems.

\end{abstract}

%%%%%%%%%%%%%%%%%%%%%%%%%%%%%%%%%%%%%%%%%%%%%%%%%%%%%%%%%%%%%%%%%%%%%%%%%%%%%%%%
\section{INTRODUCTION}
    Often the success of tactile sensing methods is judged by the accuracy or the speed with which a task can be completed, with little or no focus on the data efficiency of the methods. The requirement to collect large training sets for every new task or new sensor means these methods can be impractical and time consuming. This paper focuses on a data-efficient method of tactile sensing based on a latent variable model. 
    
    Current methods for tactile sensing try to build an explicit sensor model in a separate phase preceding the testing of the model on a task, therefore needing to train across a large data space to be representative. If any configurations are missed, or a variable is simply not accounted for, then the robot is likely to fail at its task when unexpectedly encountering these novel states during testing (e.g. in \cite{Lepora2017} changes in height were not accounted for, leading to a failure in contour following in some cases). Further problems arise when slow data collection is combined with the well-known curse of dimensionality~\cite{bellman1961}:
    the more dimensions included in the models the less representative the data becomes of all states leading to the need  for considerably more data to compensate and therefore a considerably increased  time collecting data.
    
    \begin{figure}[t]%[thpb]
    \centering
    
    \begin{tabular}[b]{@{}c@{}}
        {\bf (a) Robot Arm mounted Sensor and Stimuli}\\
        \includegraphics[width=.9\columnwidth]{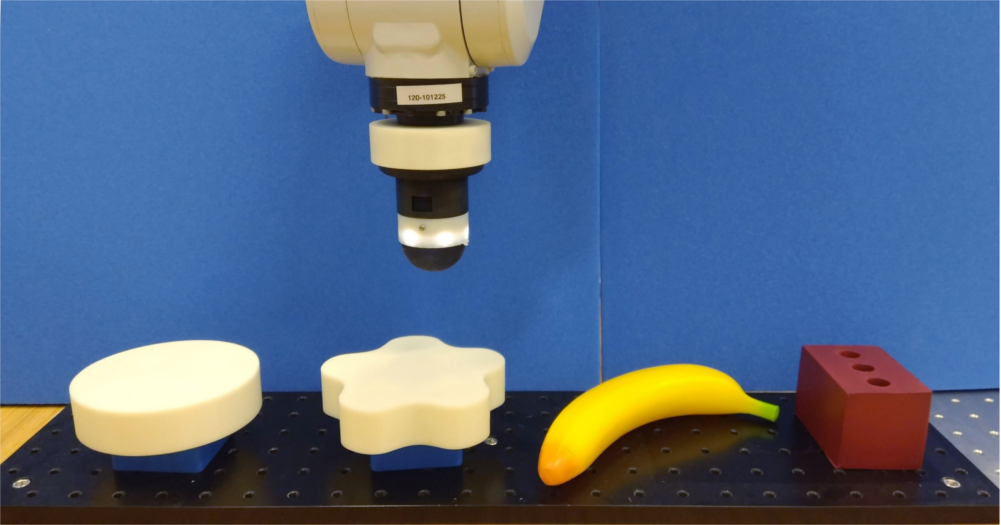}\\
        
        \begin{tabular}[b]{@{}cc@{}}
            {\bf (b) Without contact} & {\bf (c) Contacting edge}\\
            \includegraphics[width=.42\columnwidth]{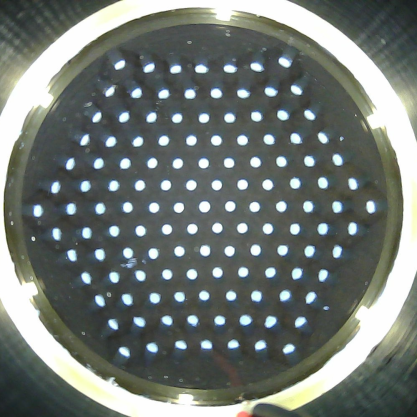} &
            \includegraphics[width=.42\columnwidth]{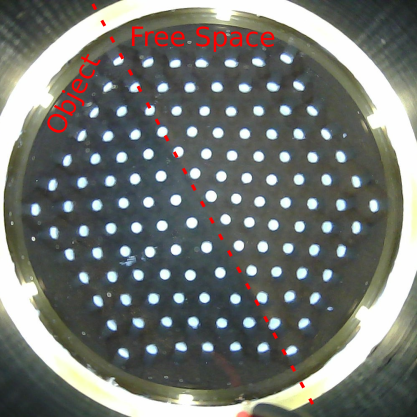}
            \end{tabular}
        \end{tabular}
    
        \caption{(a) Experimental setup showing TacTip mounted on robotic arm with circle, flower, banana and brick shaped stimuli. (b) View from TacTip internal camera showing pins in neutral position (c) View from TacTip internal camera showing pins when contacting an object edge (location indicated by the dotted line).}
        \label{robot}
        \vspace{-15pt}
    \end{figure}
    
    To overcome these problems of data efficiency this work proposes the use of a latent variable model, which needs only a single labelled data point for the initial model and uses an intelligent data collection policy to collect sufficient data throughout a task to autonomously build a  sufficiently representative model. 
    In this case the data collected will always be relevant to the task at hand, to some extent resolving the curse of dimensionality as it is no longer necessary to collect data over all possible dimensions of variation.

    With online learning, each new reading is considered for addition to the model to improve it, unlike in offline methods where the model is fixed beforehand and new data is ignored, despite the new data being representative of, and most relevant to, the task at hand. As such, this work demonstrates: 1) the novel use of data-efficient methods for tactile sensing, 2) the use of online learning in the control of a tactile sensor, and 3) the use of latent variable models in dimensionality reduction of tactile data.

%%%%%%%%%%%%%%%%%%%%%%%%%%%%%%%%%%%%%%%%%%%%%%%%%%%%%%%%%%%%%%%%%%%%%%%%%%%%%%%%
\section{Background and Related Work}
    The TacTip~\cite{chorley2009}, as used in this paper, is a biomimetic tactile sensor, consisting of a soft, deformable rubber-like gel-filled skin with 127 raised pins on the internal surface, the tips of which are tracked in 2D using an internal camera.
    
    Since its development in 2009 the TacTip has been applied to many different tasks. One of the biggest challenges is decoding the meaning of the rich sensor data, which has 254 dimensions (2D displacement of each pin), into a simple enough form to be usable in control tasks. So far all of the methods applied to this challenge are offline methods: histogram methods~\cite{Lepora2017}, Principle Component Analysis (PCA)~\cite{Aquilina2018} and deep CNNs~\cite{Lepora2018}. These methods are successful to varying degrees, with deep CNNs greatly outperforming histogram methods, but all rely on the collection of large data sets for training; specifically a set of at least 180 taps for histograms and 2000 contacts for deep CNNs.
    
    These methods have then been used to perform a variety of control tasks, including object manipulation~\cite{cramphorn2016tactile,Tian2019}, slip detection~\cite{Su2015,James2018,Veiga2015}, surface following~\cite{Yi2016,Sutanto2019} and contour following~\cite{Lepora2017,Lepora2018,martinez2013active,Aquilina2019}. Contour following is the tracking of a prominent feature, such as a well-defined object edge or ridge, often in a complete loop to discern the shape of an unknown object. This has been considered as an  example task many times with sensors such as the TacTip~\cite{Lepora2017,Lepora2018,Aquilina2019} and iCub fingertip~\cite{martinez2013active}.
    The TacTip is particularly well suited to contour following due to its ability to deform around edges. Some of the biggest challenges in contour following include the objects not being uniform in shape or height, objects being made of varying materials that deform in different ways when touched, and every day objects being highly irregular. Contour following is often done with 2D contours as this is the simplest case, with displacement in the third dimension kept constant.

    The literature concerning the combination of online learning with tactile sensing is sparse, which highlights the opportunities in this area. Previous work has focused on the use of tactile hands in object classification, using Gaussian Processes~(GPs)~\cite{soh2014incrementally, Soh2012} and Passive Aggressive online learning~\cite{Kaboli2015}, and object localisation, using particle filters~\cite{Chalon2013}.

%%%%%%%%%%%%%%%%%%%%%%%%%%%%%%%%%%%%%%%%%%%%%%%%%%%%%%%%%%%%%%%%%%%%%%%%%%%%%%%%
\section{Methods}

    \begin{figure}%[t]%[thpb]
        \centering
        \includegraphics[width=.48\columnwidth]{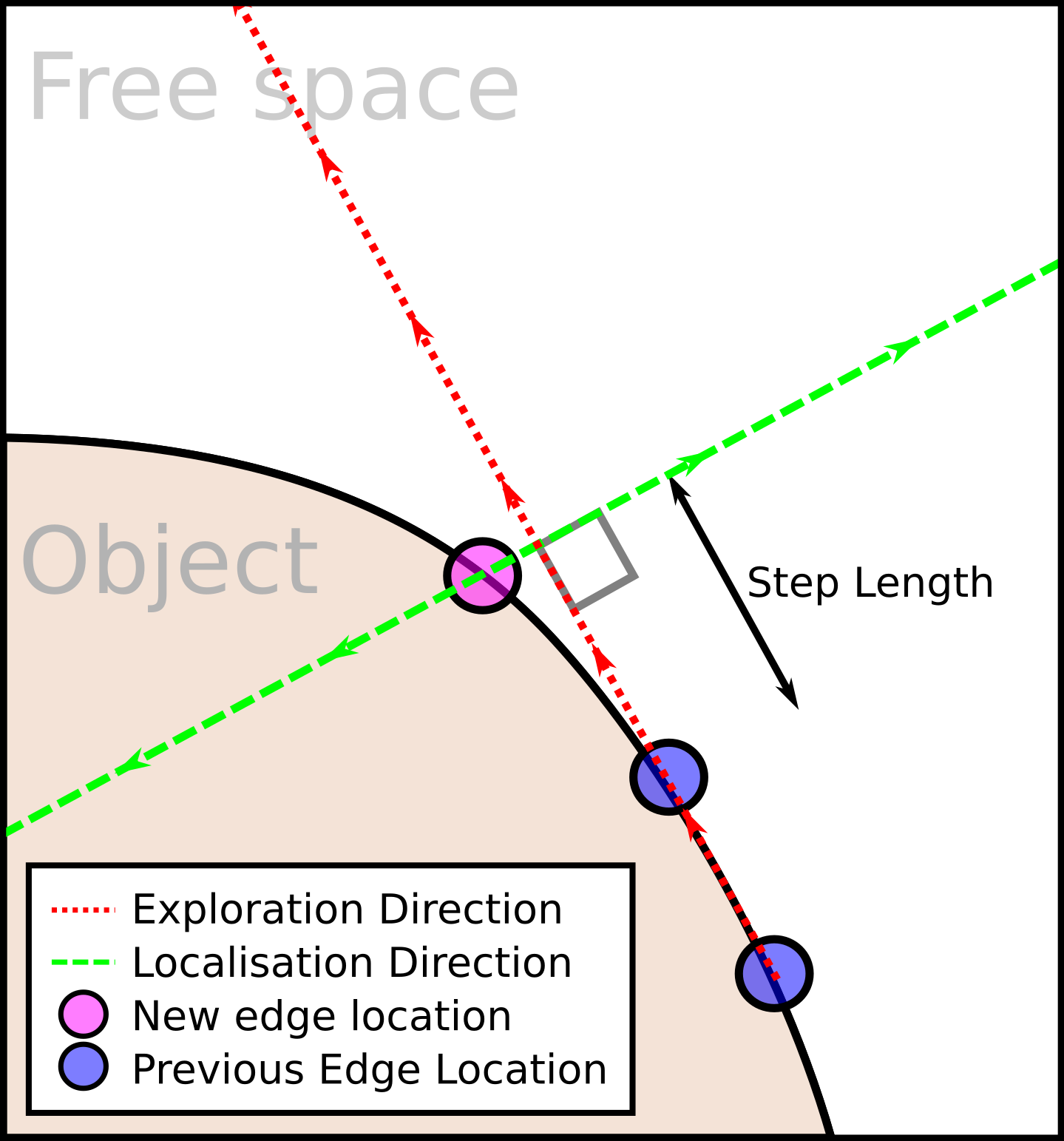}
        \includegraphics[width=.48\columnwidth]{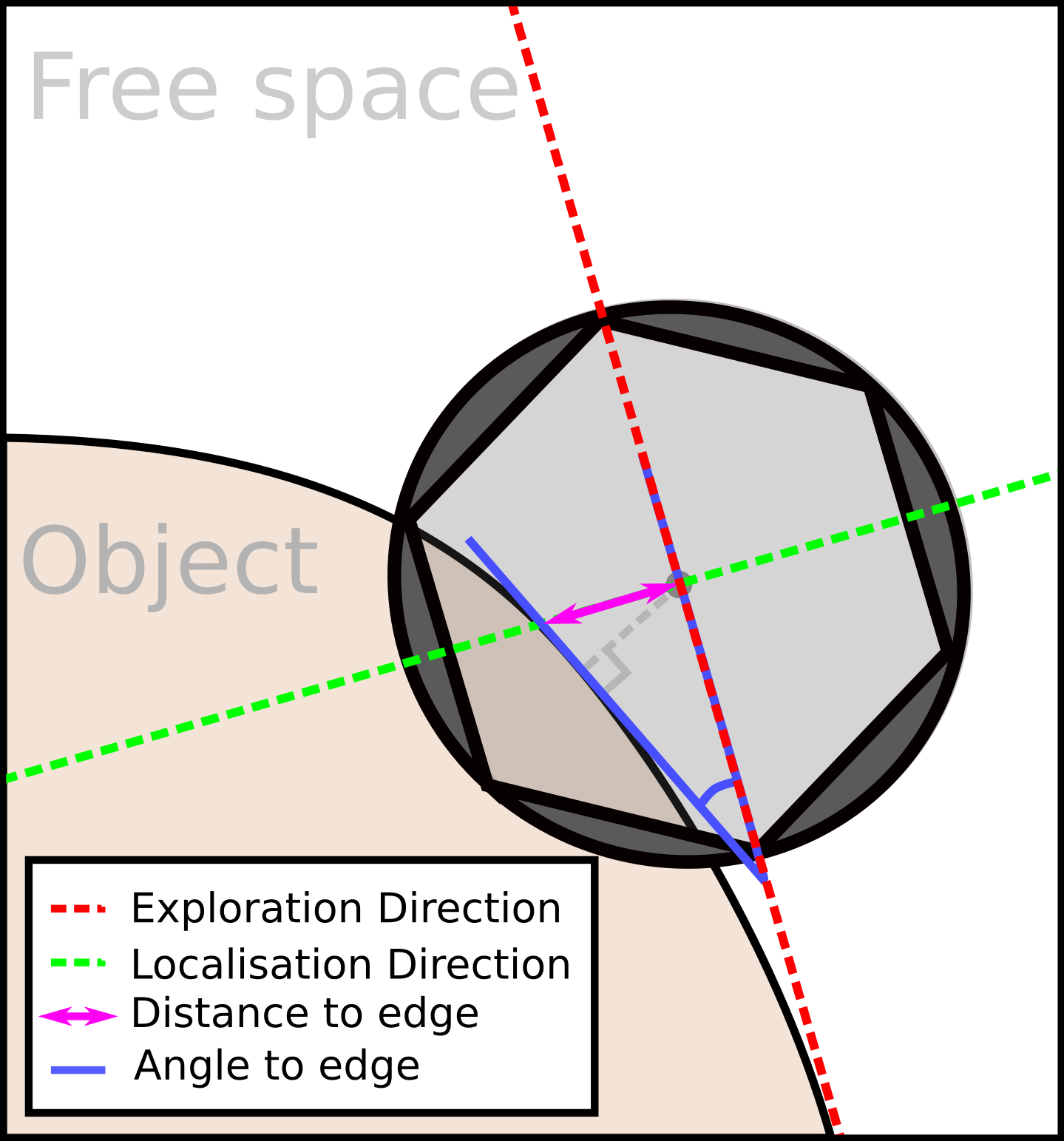}
        
        \caption{Left: How the direction of Exploration is found based on previously found edge locations. The aim of Localisation is to find the new edge location. Right: Close up of TacTip location at end of Exploration phase, where ``distance to edge" (pink) must be found during Localisation, irrespective of the orientation of the sensor (blue).}
        \label{axes}
    \end{figure}

    \subsection{Experimental Setup}
        \subsubsection{Hardware}
            A TacTip is attached to an IRB120 ABB 6-DOF robotic arm, with stimuli rigidly attached to a plate in front of the robot. This setup replicates that in~\cite{Aquilina2019}.
        
        \subsubsection{Software}
            The main algorithms are implemented in MATLAB, which is run on Windows. The robotic arm is controlled through a separate code base which also handles the detection of pin positions using computer vision. Individual pin locations are detected using OpenCV as described in~\cite{Ward-Cherrier2018}, and the average maximum pin displacement during a tap is used as the taxel data.
            
    \subsection{Algorithms}
        \subsubsection{Task Setup} 
            The task of contour following when restricted to 2D displacements, as in the case of following the edge of a planar object, is broken down into two phases. The first is Exploration which gives a prediction of the edge having moved further around the object. The second is Localisation which, having moved according to the prediction in Exploration, corrects the error in this movement (i.e. estimates the displacement from the edge perpendicular to the direction moved during Exploration). The contour following is achieved by repeating these two phases as shown in \autoref{control-policy-alg}.
            
            \begin{algorithm}
                \SetAlgoLined
                Bootstrap\;
                \While{distance to start location $>$ 1 step-length}{
                    Exploration\;
                    Localisation\;
                }
                \caption{High-level Contour Following\label{control-policy-alg}}
            \end{algorithm}
            
            This paper focuses on optimisation of the Localisation phase and therefore only a naive Exploration phase is implemented: linear extrapolation of the two previously found edge locations, as shown in \autoref{axes} and \autoref{exploration-alg}. 
            
            \begin{algorithm}
                \SetAlgoLined
                Extrapolate in a straight line between the locations of the previous two least dissimilar points (call this the Exploration line)\;
                Rotate the sensor to maintain alignment with Exploration line\;
                Move along Exploration line by step length\;

                \caption{Exploration\label{exploration-alg}}
            \end{algorithm}
            
        \subsubsection{GP-LVM}
            To efficiently achieve Localisation, accurate estimation of the displacement of the sensor, $r$, from the edge along any axis, irrespective of any other variables, is required given only the taxel pattern, $y$, from a tap. Hence a Gaussian Process Latent Variable Model (GP-LVM)~\cite{Lawrence2004} is used to predict the displacement of the edge from the sensor irrespective of any other variables. The orientation of the sensor is not estimated directly, but a proxy for angle and all other variations (e.g. edge height, compliance or sharpness), termed $\phi$ is inferred as part of the method; this variable is the latent variable or unlabelled variable in the problem. The input to the GP-LVM is thus $(r, \phi)$ and the output is $y$.
            
            \begin{algorithm}[t]
                \SetAlgoLined
            
                Tap and estimate the distance to the edge, along the Localisation line (perpendicular to the Exploration line), using GP-LVM\;
                Move the estimated distance along the Localisation line\;
                Tap again and estimate the new distance to the edge\;
                
                    \If{distance to edge $>$ tolerance (i.e. 2mm)}{
                        Collect a series of evenly spaced taps along the Localisation line\;
                        Find displacement of taps from the edge using the dissimilarity measure\;
                        Optimise $\phi$ of taps\;
                        Add this labelled data to the model\;
                    }
                
                \caption{Localisation\label{localisation-alg}}
            \end{algorithm}
            
            A GP-LVM is similar to a Gaussian Process (GP)~\cite{Rasmussen2006} in that it is built on the assumption that outputs (observations) follow a multi-variate Gaussian distribution. It differs in that a GP is used to calculate a novel output, $y$ (i.e. taxel pattern), given a novel input, $x:=(r, \phi)$ (i.e. displacement and orientation), whereas a GP-LVM does the opposite, inferring the input (the latent variables) given a novel output. This is achieved via optimisation of the $D \times n$ input using a set of known input-output pairs (the finding of such pairs is described later). The objective function to optimise is given by
            \begin{equation} \label{max_log_marg_gplvm}
                L = -\frac{D}{2}\trace\left(K^{-1} \frac{1}{D} y y^T\right) - \frac{D}{2}\log|K| - \frac{D n}{2}\log 2\pi
            \end{equation}
            which is the log marginal  likelihood~\cite{Lawrence2004} for GP-LVMs. Here $K$ is the covariance matrix where the covariance function, $k$, is the squared exponential function~\cite{Rasmussen2006}
            \begin{equation} \label{cov_func_sqrex_multidim}
                k(\bm{x},\bm{x'}) = \sigma_f^2\,\exp\left[-\left(\frac{(r-r')^2}{2l_r^2}+\frac{(\phi-\phi')^2}{2l_\phi^2}\right)\right].
            \end{equation}
            This depends on the hyperparameters $\theta := (\sigma_f, l_{r}, l_{\phi})$ --- the variance of the (noise free) signal,  characteristic length-scale of displacement and characteristic length-scale of $\phi$ respectively --- which  must be optimised for this problem. 
            The signal noise is represented by another parameter, $\sigma_n$, in $K$ and must also be tuned to the problem, but this value is assumed problem independent for a given setup as it represents the system noise (e.g. sensor noise and robot positioning noise). It was calculated from representative data beforehand to be $\sigma_n=$1.14.
            
            Note that the value of (\ref{max_log_marg_gplvm}) is  dependent on both the inputs, $x$, and the outputs, $y$. The GP-LVM can then therefore be used to infer the novel input $x$ given a novel output $y$. 
            
            The model consists of both optimised hyperparameters $\theta$ and known input-output pairs $(x,y)$ representative of the data; when data is collected during testing, the set of input-output pairs is augmented. Hence, the accuracy of the model relies on the accuracy of the automatic labelling of new data. In the case of $r$, the true (relative) displacements between taps on the same Localisation line are known, so the only inaccuracy in $r$ occurs in identifying the offset from the edge. The true value of $\phi$ is not known (being the representation of all variation not caused by $r$), however it is assumed that $\phi$ is constant along any given Localisation line and all variation in the output along that line is driven by $r$.
            
            As the optimisation of the log marginal  likelihood requires calculating the inverse of $K$, the complexity is $\mathcal{O}(n^3)$. Therefore the quantity of data in the model is kept to a minimum by not adding every tap taken during an experiment. Instead, only taps collected according to the following data collection policies are added.

        \subsubsection{Dissimilarity}
            In the beginning only a single labelled data-point is given, defining the desired state of the sensor during contact with an edge (named the ``reference tap"). A way of labelling new data collected at unknown displacements and orientations is therefore needed to add useful information to the model. To accomplish labelling of displacement we make use of a dissimilarity measure, namely the Euclidean distance
            \begin{equation} \label{euclid-equation}
                d_{Euc}(\mathbf{a},\mathbf{b}) = \sqrt{\sum_{i=1}^{n}(a_i-b_i)^2} 
            \end{equation}
            to quantify the dissimilarity between unlabelled data and the reference tap.
            
            Using this measure the edge of the object is identified as the point with the smallest dissimilarity with the reference tap. The Euclidean distance is scalar so the direction of the edge from a single tap cannot be found, but with multiple taps in a straight line the dissimilarity profile will have a clear minimum where the edge is encountered (\autoref{dissim-profile}).
        
        \subsubsection{Online Policies}
            \begin{figure}%[t]
                \centering
                
                \includegraphics[width=\columnwidth]{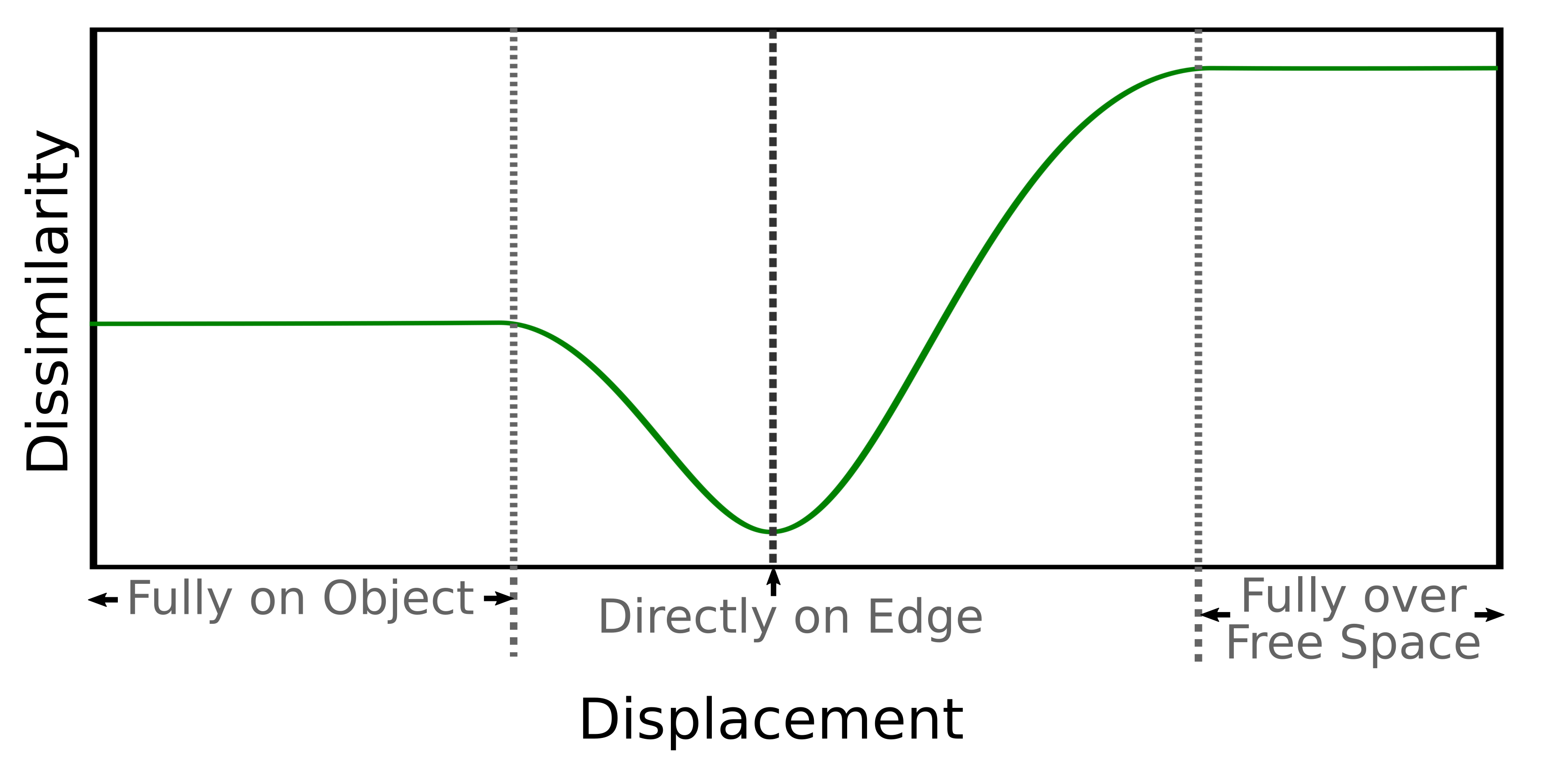}
                
                \caption{Diagram of how dissimilarity should change with displacement from an edge, given that the reference tap was taken directly on the edge and assuming the object is flat topped with uniform height. }
                \label{dissim-profile}
            \end{figure}
        
            To build the GP-LVM model online a data collection policy is defined, capable of collecting relevant data autonomously.  As the aim is to be efficient in terms of number of taps, new data is only collected and added to the model when the model is inaccurate. This inaccuracy was assessed by doing two taps; one at the intersection of the Exploration line and the Localisation line, and one after moving the predicted displacement. This second tap should be directly on the edge if the model is accurate, giving a new prediction of $0\,\text{mm}$ to the edge (within a tolerance -- here $2\,\text{mm}$ was used), otherwise the model is giving contradictory predictions and is therefore inaccurate. 

            Having found the model is inaccurate, new data needs to be collected and added to the model to improve the predictions, as it is assumed the current data in this region is anomalous or non-existent.
            The new data consists of an evenly spaced line of taps along the Localisation line between $\pm10\,\text{mm}$ from the intersect with the Exploration line.
            
            By finding the dissimilarity for these taps, which were taken at known intervals of displacement,  the dissimilarity profile is found which will have a distinct minimum at the displacement where the edge is (i.e. in the location most similar to the edge), as shown in \autoref{dissim-profile}. The location of the minima can be estimated with greater resolution than the displacement interval through use of a separate GP trained only on the new data. Knowing the relative location of this minima enables the real displacements from the edge to be calculated for this line of taps.
            
            The value of the latent variable $\phi$ for this particular Localisation line is inferred using the GP-LVM. The data was collected with the sensor at the same orientation to the edge, assuming all other variables are constant. The value of $\phi$ is therefore  the same for the entire line of taps, meaning only a single value needs inferring. This approach shares similarities with the alignment learning method proposed in~\cite{kazlauskaite2019}, however that method has been applied only in an abstract setting and does not exploit all the structure of this problem.
            
            Obviously this procedure cannot start on its own as two previous edge points must have been collected for the initial Exploration step. An initialisation phase, termed Bootstrap, is used to collect the first two edge locations. This involves the steps outlined in \autoref{bootstrap-alg}.
                
            This leads to the full control policy described by Algorithms \ref{control-policy-alg}, \ref{exploration-alg}, \ref{localisation-alg},  and \ref{bootstrap-alg}.

            \begin{algorithm}
                \SetAlgoLined
            
                Collect 3 taps in a right-angled triangle\;
                Calculate the dissimilarity to the reference tap for all 3 taps\;
                Calculate the direction of greatest decrease in dissimilarity\;
                Tap along this direction collecting evenly spaced taps and label (in the same fashion as in \autoref{localisation-alg})\;
                Identify and move to edge location\;
                Move perpendicular to direction of greatest decrease in dissimilarity by one step length\;
                Localisation (\autoref{localisation-alg})\;
            
                \caption{Bootstrap\label{bootstrap-alg}}
            \end{algorithm}

%%%%%%%%%%%%%%%%%%%%%%%%%%%%%%%%%%%%%%%%%%%%%%%%%%%%%%%%%%%%%%%%%%%%%%%%%%%%%%%%
\section{Results}
    \subsection{Offline Testing}
        Offline testing was done to evaluate the methods on a representative dataset. Having been given only 5 lines of taps to build the model, simulating data collection at 5 different angles, GP-LVM was able to predict the displacement and other variations, $\phi$, of a dataset of taps taken over a range of displacements (-10 to +$10\,\text{mm}$) and angles (-45\textdegree to 45\textdegree) from an edge. In this case $\phi$ is taken as a direct proxy of angle as no other variables change. Here $\phi$ is ranged between -2 and 2, corresponding to -45\textdegree and 45\textdegree respectively. The predictions are good, except at the extremes of displacement where the $\phi$ predictions converge towards 0; this is expected because here the sensor is fully on a flat surface or fully over free space, both of which are invariant to rotation.
        
        Even with fewer training lines the error in displacement predictions are small, with the highest mean error being $1.02\,\text{mm}$ when only one training line (and therefore only one orientation) was given. Even this precision is enough to complete contour following, although the standard deviation is relatively large meaning this accuracy may not always be maintained. 
        
        These results indicate that the use of a single training line should be sufficient for task completion, therefore the online methods were designed to collect only a single line during start up. When this is no longer sufficient more data is collected automatically, using the algorithm as described.
        
        \begin{table}[t]
            \caption{Table of offline results. An anomaly is defined as a $r$ prediction with magnitude greater than $12\,\text{mm}$ (as the maximum should be $10\,\text{mm}$), and for $\phi$ predictions of magnitude greater than 3 (maximum should be 2).}
            \begin{tabulary}{\columnwidth}{@{}|C||RR|RR|CC|@{}}
                \hline
                \multirow{3}{26pt}{No. of training lines} 
                & \multicolumn{2}{c|}{Error in $r$ (mm)} & \multicolumn{2}{c|}{Error in $\phi$} & \multicolumn{2}{c|}{No. of Excluded } \\
                & & & & & Anomalies & \\
                 & Mean & SD & Mean & SD   & $r$  &  $\phi$\\
                \hline
                1 &     1.02  &     1.56 &     1.05  &     1.29 & 7 & 3\\
                3 &   0.48 &     0.70 &     0.17 &      0.32 & 1 & 6 \\
                5 & 0.48 & 0.62 & 0.13 & 0.21 & 2 & 2\\
                \hline
            \end{tabulary}
            \label{offline-results-table}
        \end{table}
        
        \begin{figure}[bh] 
            \centering
            
            \begin{tabular}[b]{@{}cc@{}}
                {\bf \small Step Length = 5mm} & {\bf \small Step Length = 10mm} \\[0pt]
                \includegraphics[width=.47\columnwidth]{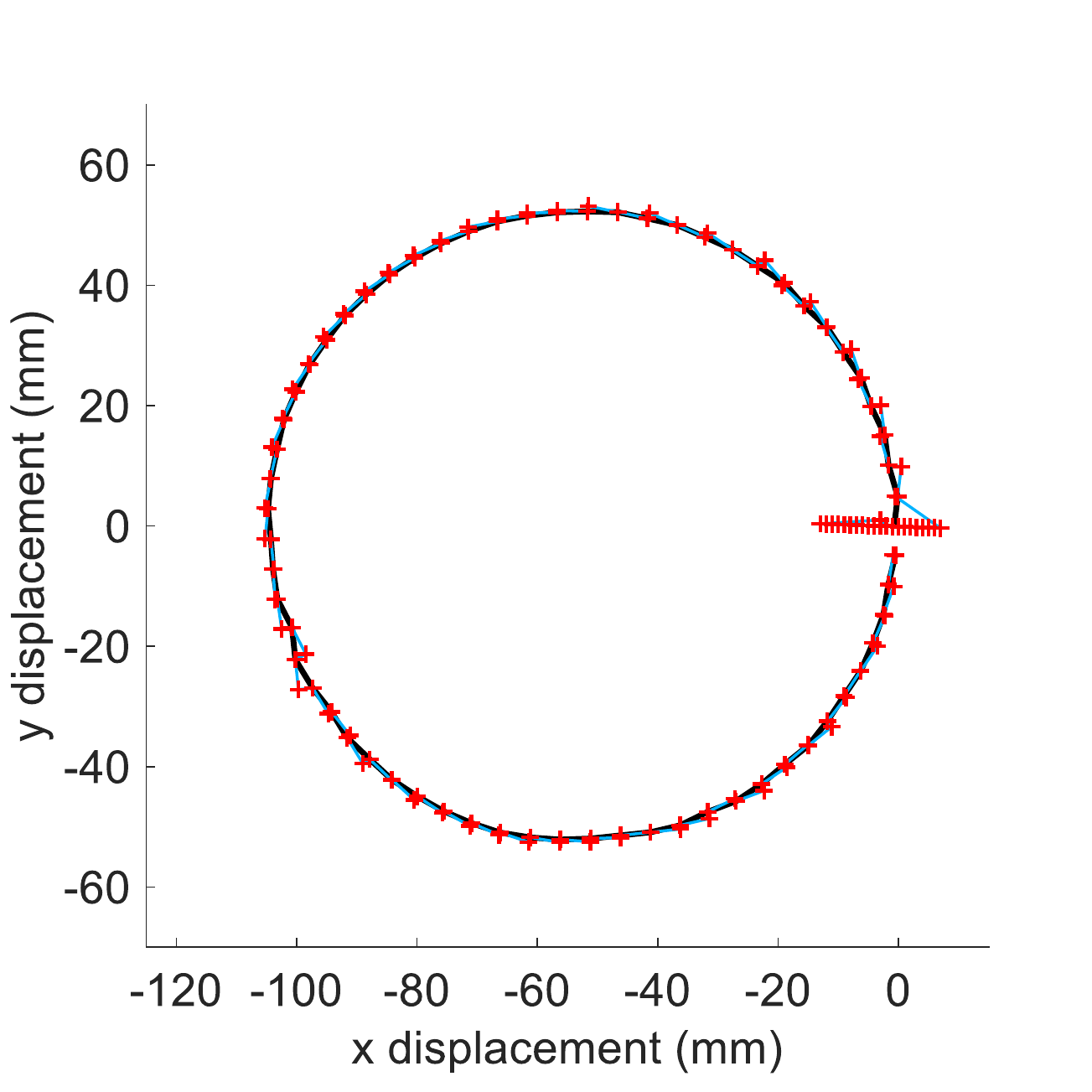} &
                \includegraphics[width=.47\columnwidth]{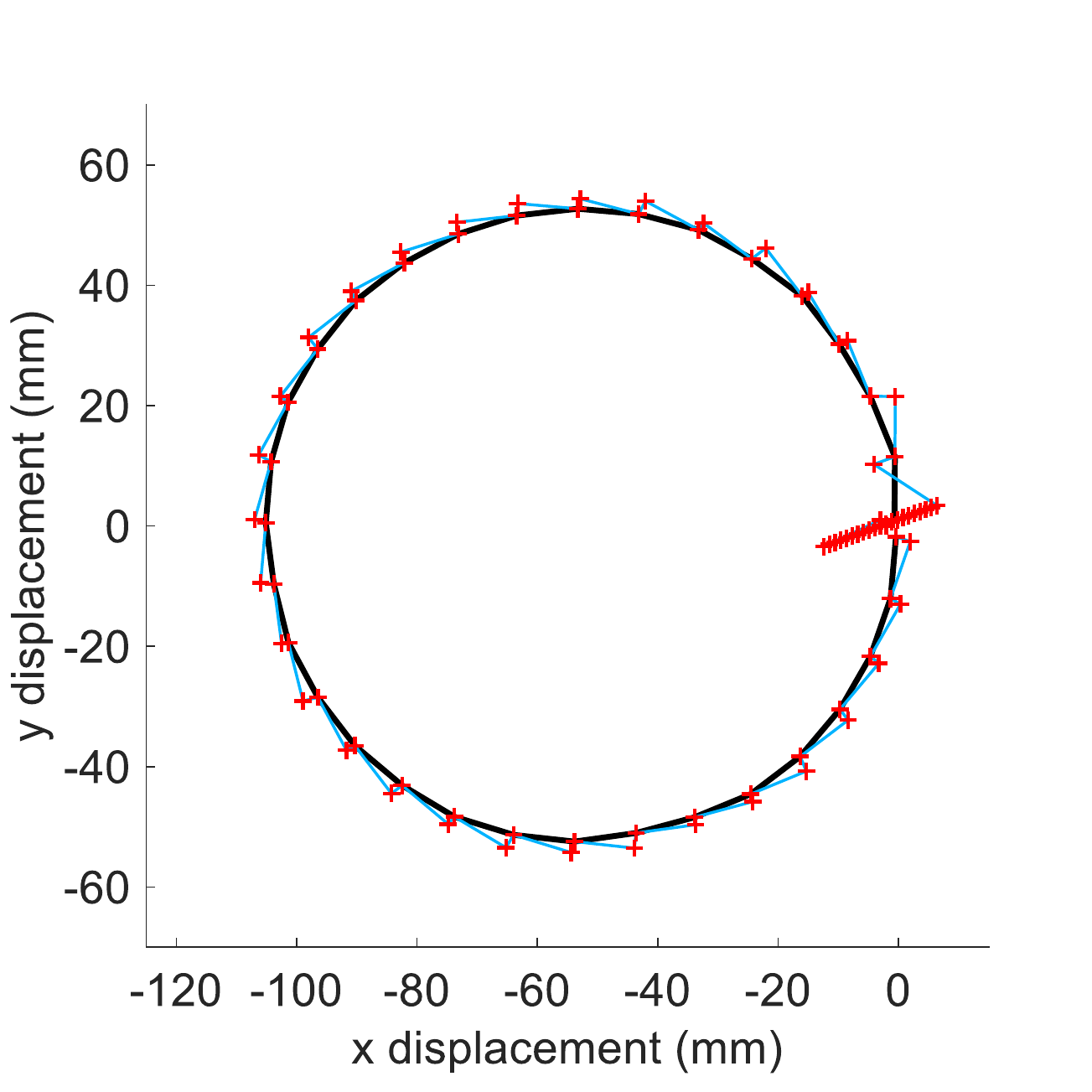} \\[5pt]
                {\bf \small Step Length = 15mm} & {\bf \small Step Length = 20mm}\\[0pt]
                \includegraphics[width=.47\columnwidth]{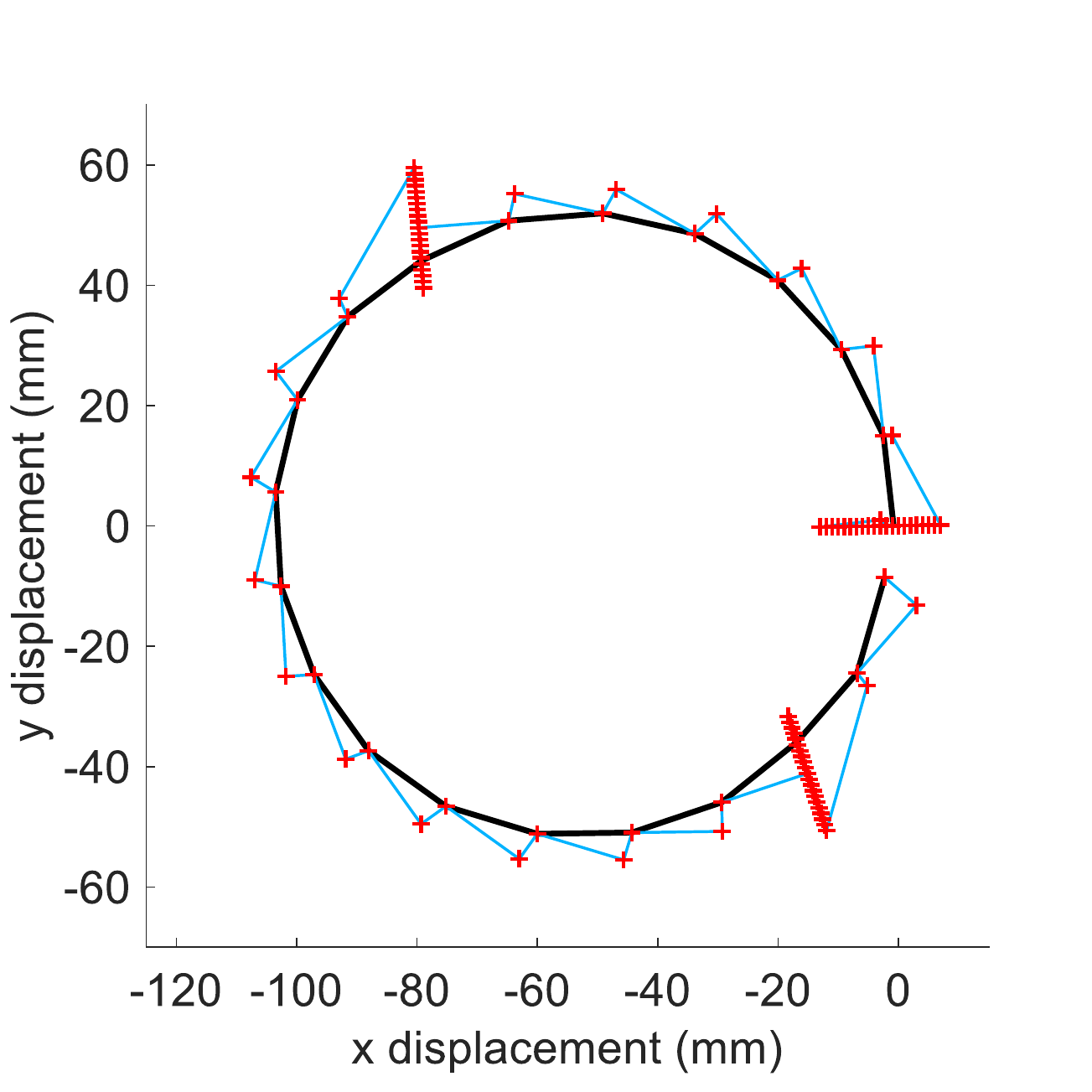} &
                \includegraphics[width=.47\columnwidth]{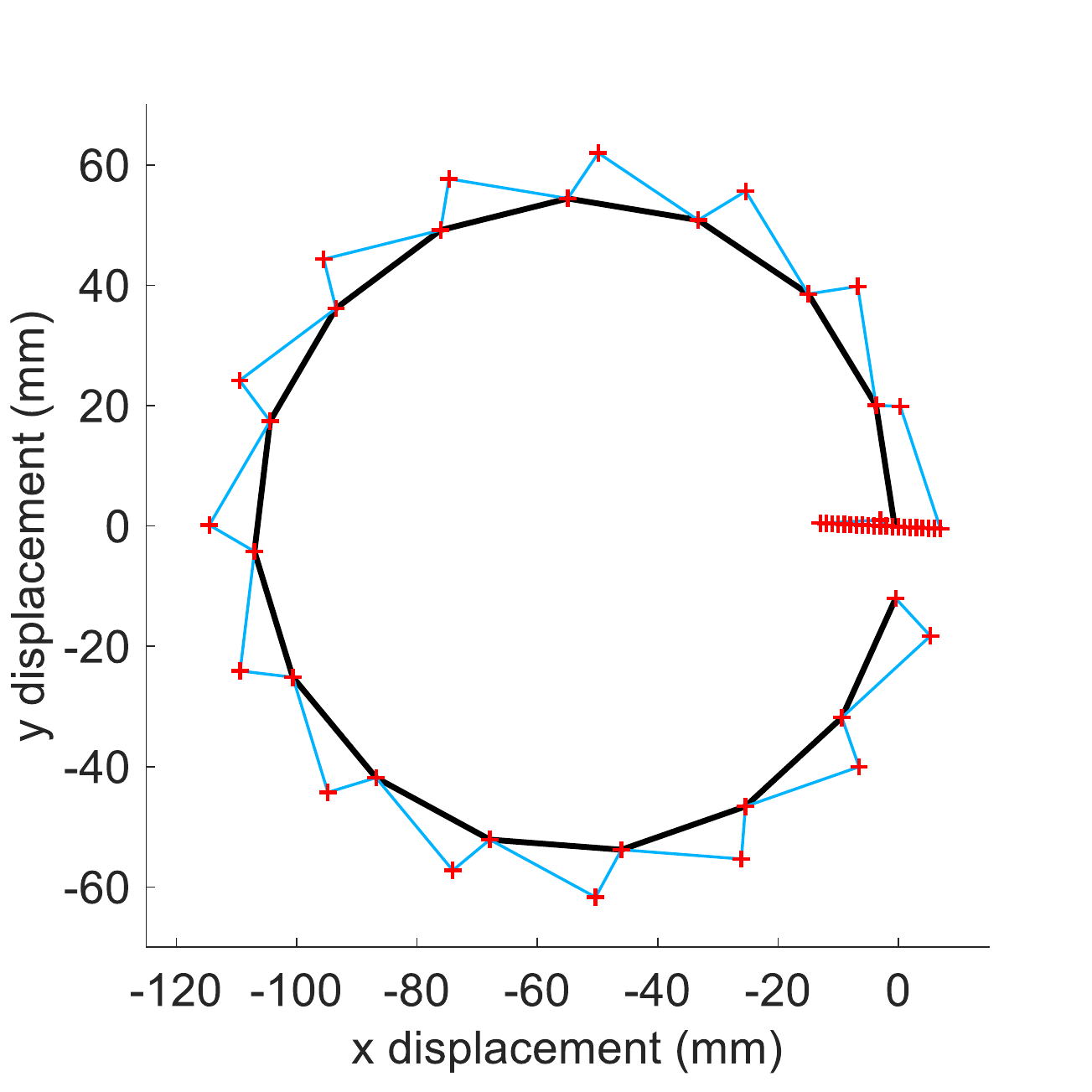}
            \end{tabular}
            
            \caption{Varying step length with the circle stimulus. Red crosses are locations where the robot tapped, blue lines show the movement between taps, and the black line joins the locations identified as the object edge. Start location was (0,0) and movement is anticlockwise.}
            \label{step-circle}
        \end{figure}
        
    \subsection{Online Testing}
        The online system was tested on a task of contour following around a circle and a flower-like stimuli. The circle is the simplest object as the edge is uniform in curvature. The flower is designed to test the system on varying edge curvatures, with both positive and negative curvatures.
        
        Given only one line of 21 taps, at $1\,\text{mm}$ intervals, the robot was able to complete the task on both stimuli, tracking the object edge until the start location was successfully reached again. Following this success the system was pushed to its limits with the following experiments.
    
        \begin{figure}[t]
            \centering
            
            \begin{tabular}[b]{@{}cc@{}}
                {\bf \small Step Length = 5mm} & {\bf \small Step Length = 10mm} \\[0pt]
                \includegraphics[width=.47\columnwidth]{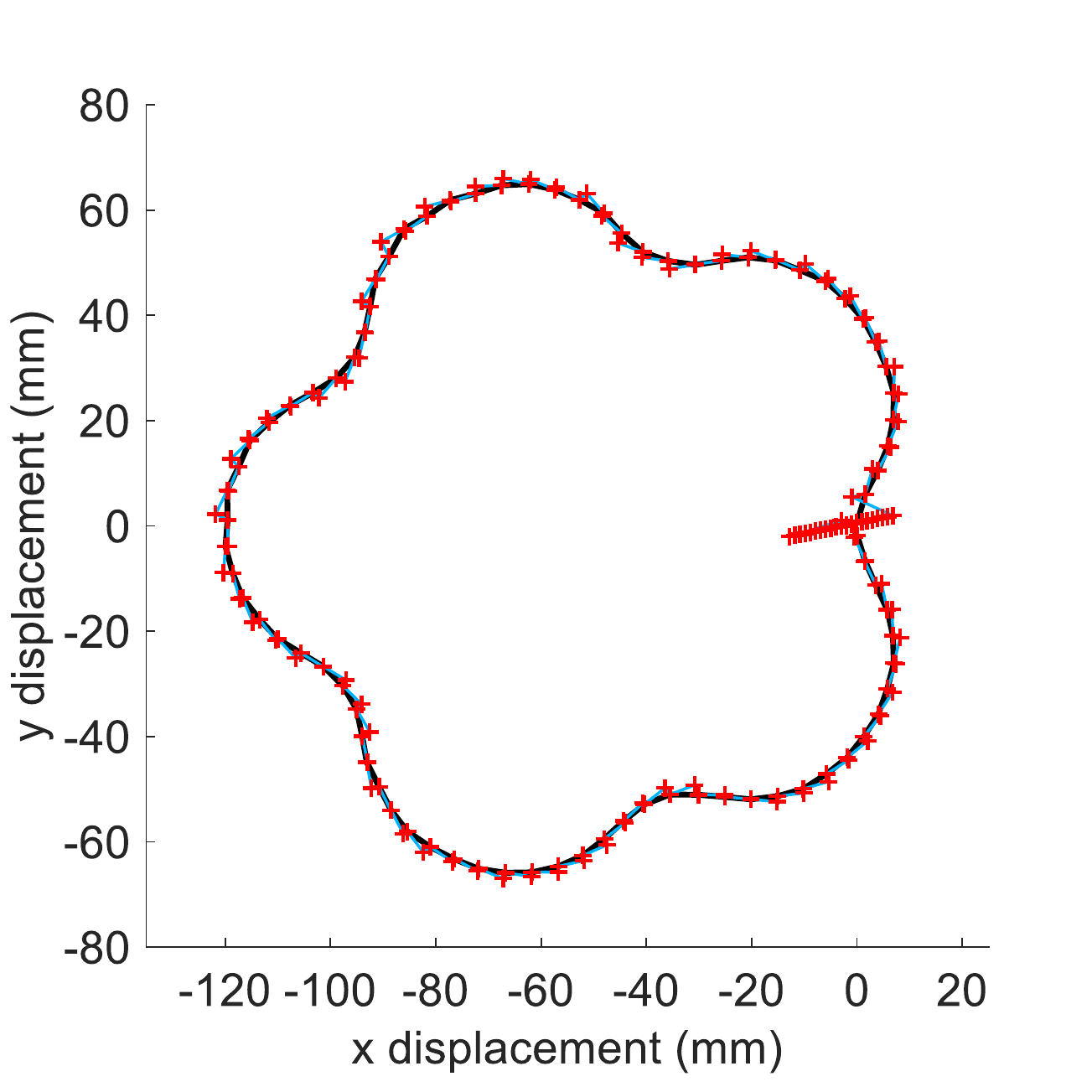} &
                \includegraphics[width=.47\columnwidth]{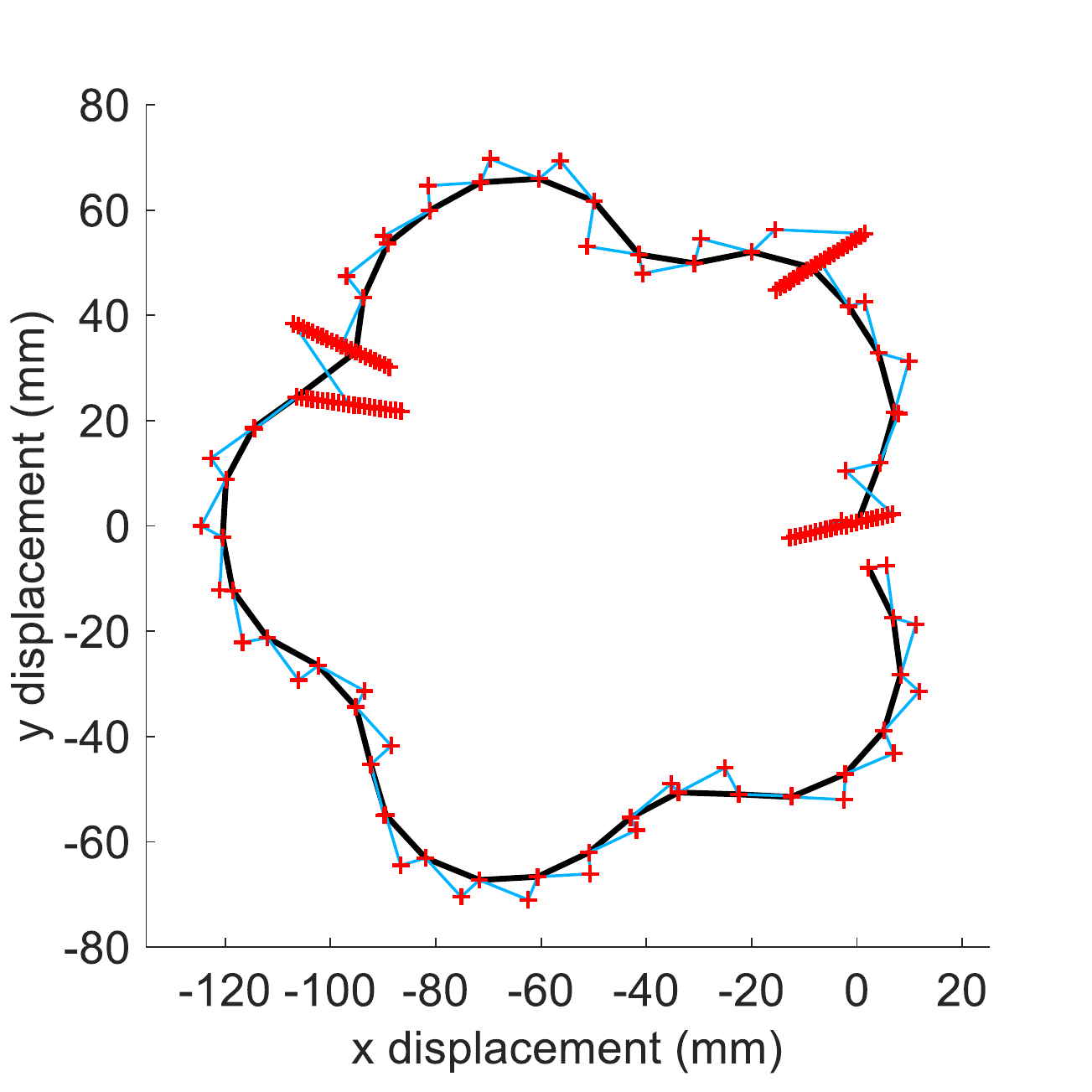}
            \end{tabular}

            \caption{Varying step length with flower stimulus. Red crosses are locations where the robot tapped, blue lines show the movement between taps, and the black line joins the locations identified as the object edge. Start location was (0,0) and movement is anticlockwise.}
            \label{step-flower}
            \vspace{-5pt}
        \end{figure}
    
        \subsubsection{Robustness to Larger Edge Displacements}
            By increasing the step length during exploration, the greater distance the edge is likely to have deviated from the Exploration line, testing the capabilities of the Localisation (which corrects these perturbations) over a larger range. With the circle stimulus the step length was increased from $5\,\text{mm}$ in steps of $5\,\text{mm}$ until $20\,\text{mm}$. All of these trials were successful using only a single line of data except for $15\,\text{mm}$ which needed 2 extra lines of data (automatically collected by the algorithm), as shown in \autoref{step-circle}. The shape of the circle is clearly visible in all cases, despite the $20\,\text{mm}$ case needing very few Exploration steps to complete the task. As can be seen in \autoref{online-results-table-steps} the error in estimation of the location of the circle edge is very small, with the smallest mean distance being $0.47\,\text{mm}$ and the greatest being $1.33\,\text{mm}$. 
            
            \begin{table}[t]
                \centering
                \caption{Mean and Standard Deviation (SD) of the shortest distance between actual circle and predicted circle location with varying Exploration step length. }
                \begin{tabular}{@{}|c||cc|@{}}
                    \hline
                    Step Length (mm) & Mean distance (mm) & SD (mm)\\
                    \hline
                    5 &     0.72 &     0.39\\
                    10  & 0.47 & 0.42\\
                    15 & 1.33 & 0.63\\
                    20  & 1.07 & 0.53\\
                    \hline
                \end{tabular}
                \label{online-results-table-steps}
            \end{table}
        
            The same was attempted with the flower stimulus but owing to the tighter curves of the ``petals", the step length could not be increased as much. The results can be seen in \autoref{step-flower}, where it is obvious the large changes in edge direction compared with the step length caused the system to collect more data as necessary; 4 lines of data were needed to build a sufficient model to complete the task and the flower shape is clearly visible.
        
        \subsubsection{Robustness to Sparse Data Collection}
            \begin{figure}%[hb]
                \centering
                \vspace{2pt}
                {\bf \small 21 Taps per Data Collection Line}
                
                  \includegraphics[width=.49\columnwidth]{pics/prettier_step5_circ_pts21_svg-tex.pdf}
                  \includegraphics[width=.49\columnwidth]{pics/prettier_step5_flower_pts21_svg-tex.pdf}

                {\bf \small 11 Taps per Data Collection Line}
                  
                  \includegraphics[width=.49\columnwidth]{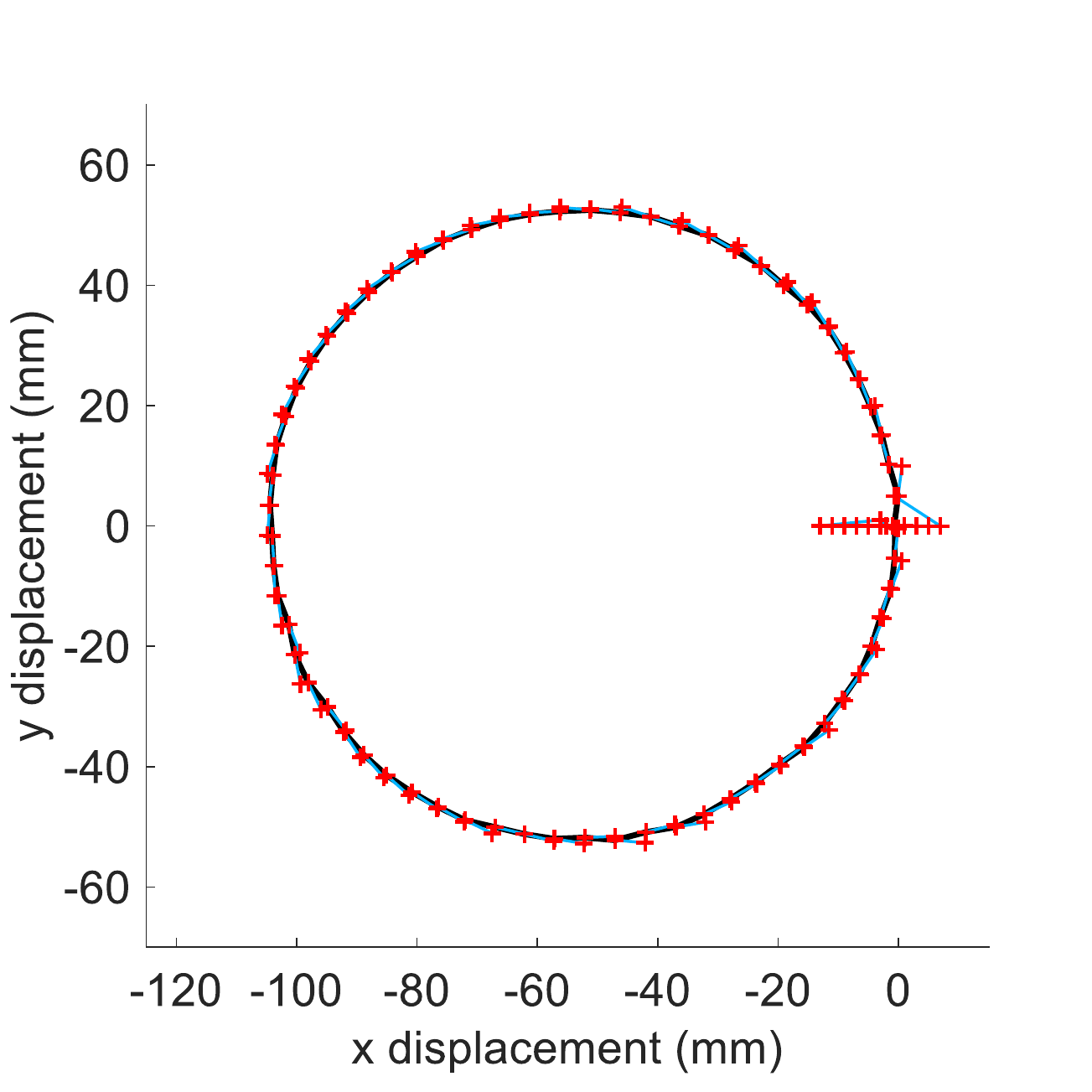}
                  \includegraphics[width=.49\columnwidth]{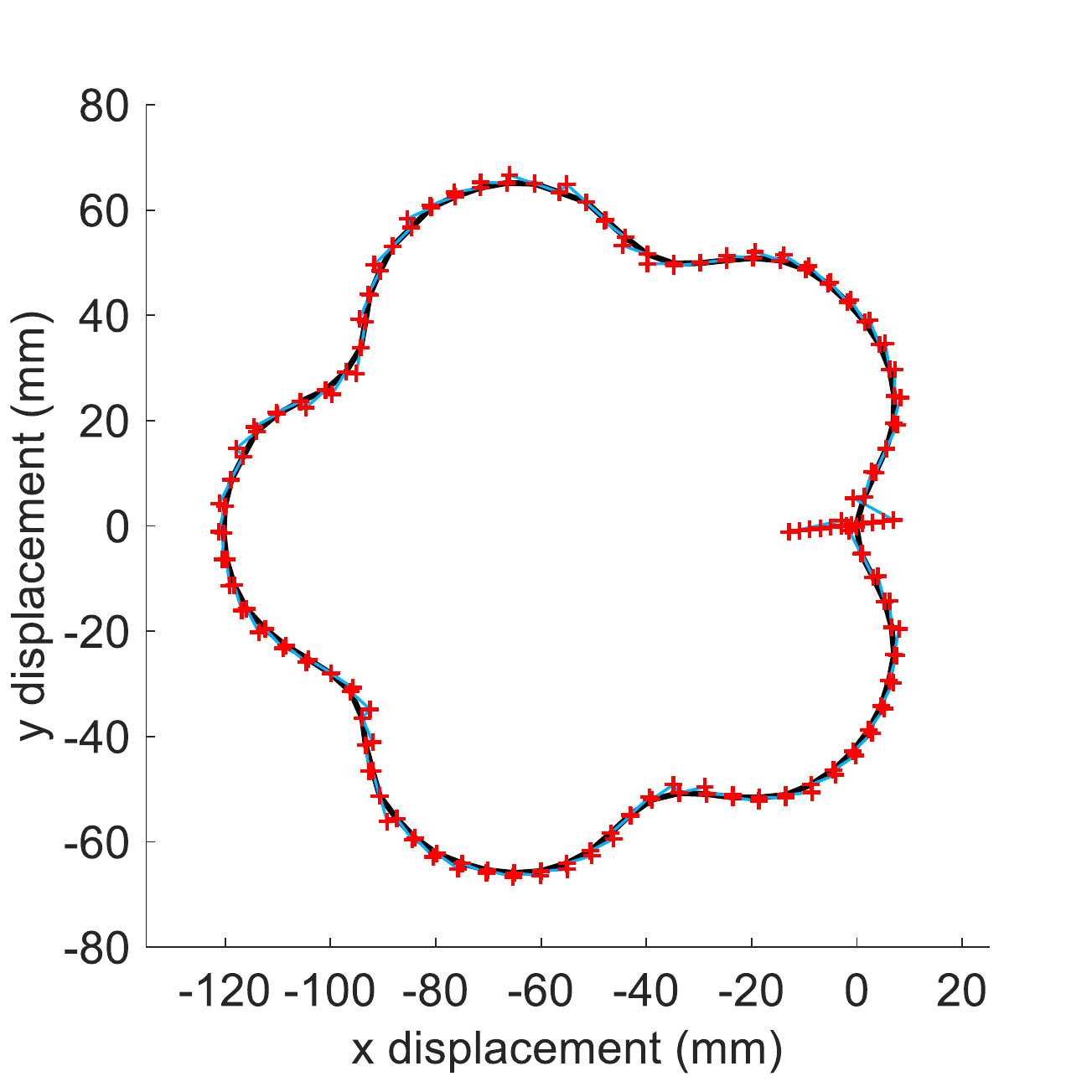}

                {\bf \small 6 Taps per Data Collection Line}
                
                \includegraphics[width=.49\columnwidth]{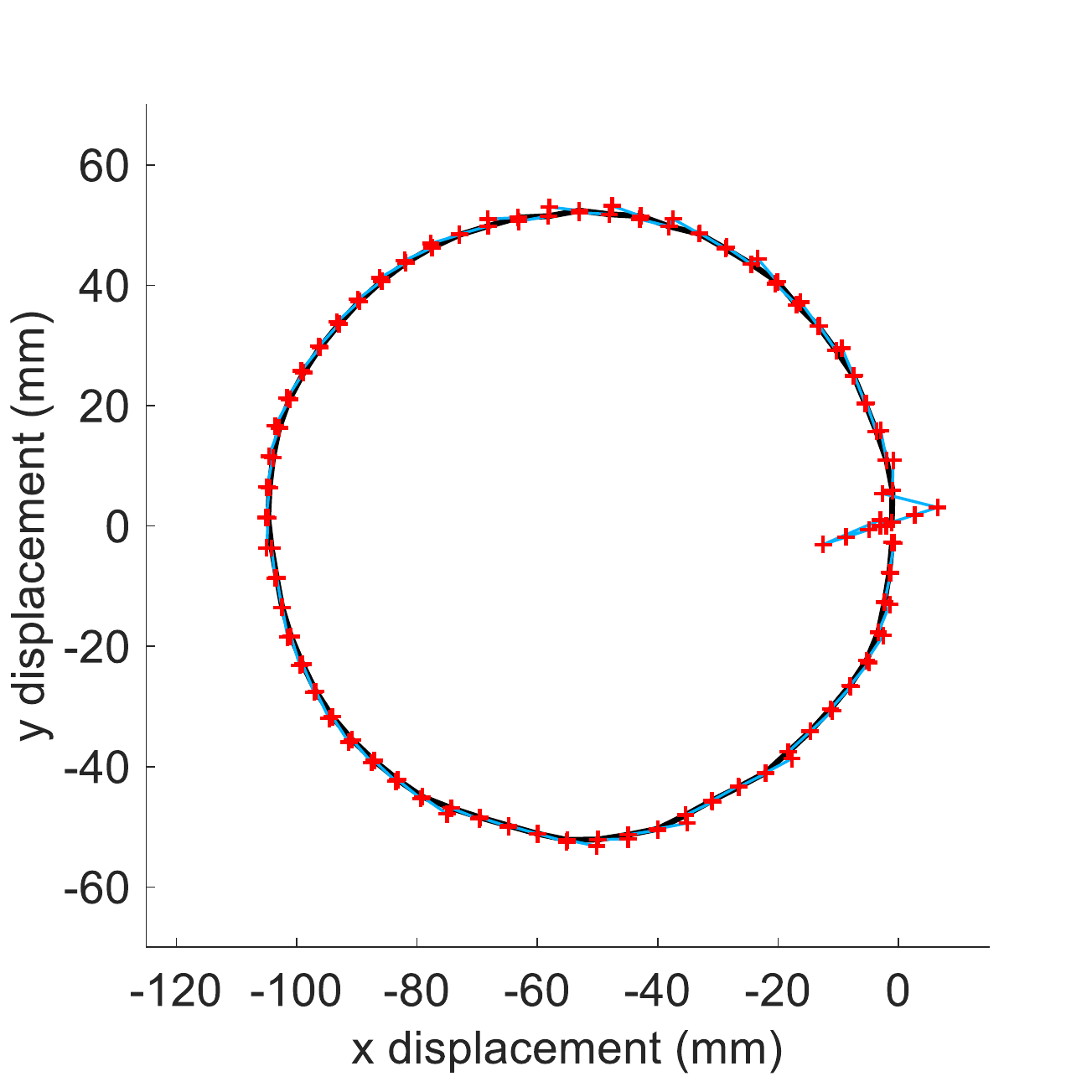}
                \includegraphics[width=.49\columnwidth]{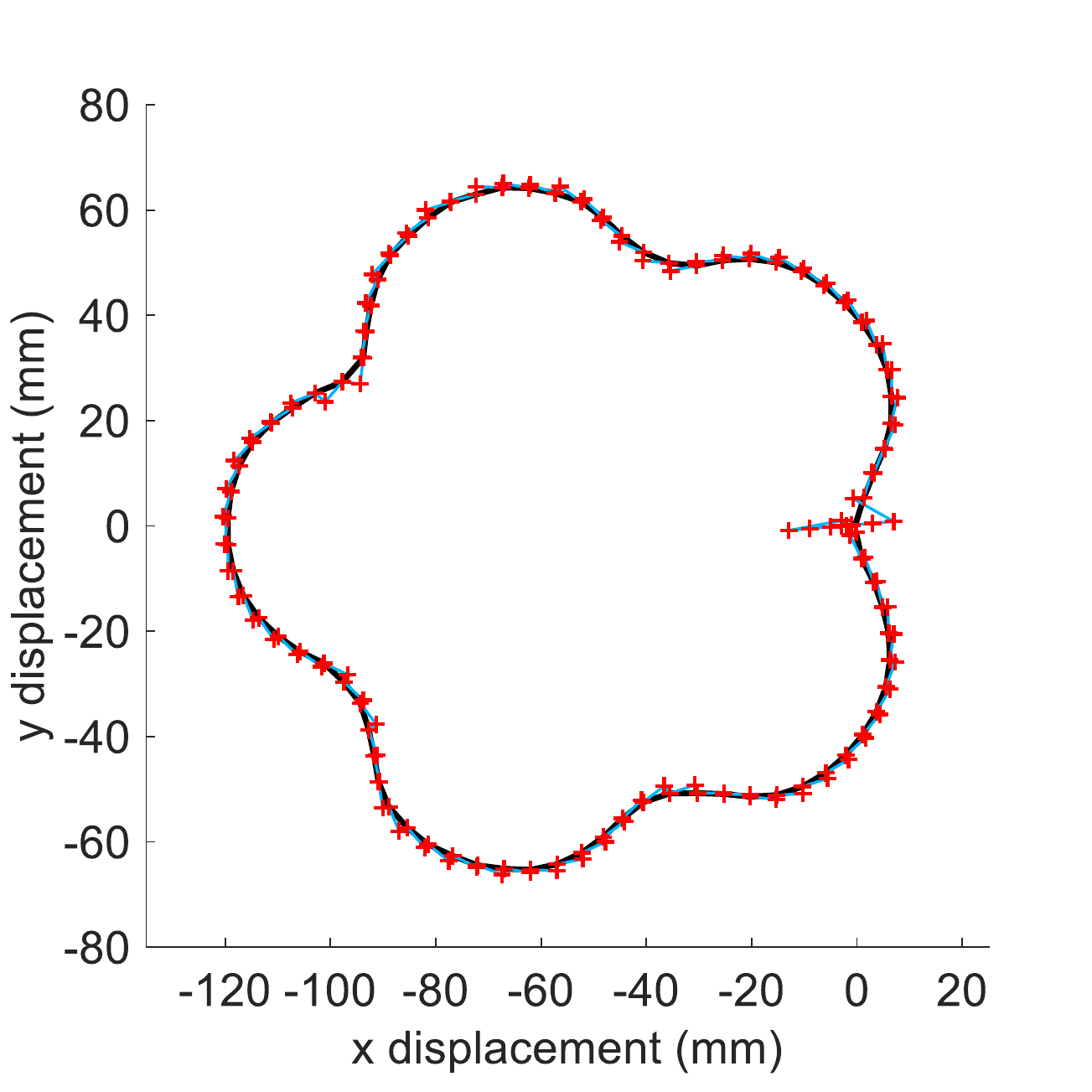}

                \caption{Using only 6 taps per data collection line on the flower and circle stimuli. Red crosses are locations where the robot tapped, blue lines show the movement between taps, and the black line joins the locations identified as the object edge. Start location was (0,0) and movement is anticlockwise.}
                \label{points-flower}
                \vspace{-5pt}
            \end{figure}
            
            \begin{table}
                \centering
                \caption{Mean and Standard Deviation of the shortest distance between actual circle and predicted circle location with varying number of taps per data collection line. }
                \begin{tabulary}{\columnwidth}{@{}|C||cc|@{}}
                    \hline
                    Number of taps & Mean distance (mm) & SD (mm)\\
                    \hline
                    21 &     0.72 &     0.39\\
                    11  & 0.74 & 0.42\\
                    6 & 1.10 & 0.47\\
                    \hline
                \end{tabulary}
                \label{online-results-table-taps}
            \end{table}
            
            The main aim of this system is to reduce the quantity of data needed to complete the task. By reducing the number of samples along the localisation line when collecting tap data to add to the model, the number of taps can be significantly reduced. To test this the distance between taps was doubled from $1\,\text{mm}$ to $2\,\text{mm}$ to $4\,\text{mm}$, giving a total of 21, 11 and 6 taps per line respectively. This should be easily handled due to the use of GP regression for estimating the minimum dissimilarity, and the use of GP-LVM for estimating displacement and $\phi$. In all three cases the task was successfully completed. Even with the use of only 6 taps per line, the edge was closely tracked with only 1 line of data on both the circle and the flower, as shown in \autoref{points-flower}. For the circle, it can be seen in \autoref{online-results-table-taps} that even when the number of taps per data collection line is reduced to 6 the location error is only $1.10\,\text{mm}$, only $0.38\,\text{mm}$ greater than with 21 taps. 
        
        \subsubsection{Robustness to Every-day Objects}\label{natural-results}
            The final test was using every-day objects which better reflect the irregularity of objects outside a laboratory setting. The two objects chosen were the foam brick and plastic banana from the YCB dataset (items ID60 and ID11 \cite{Berk2015} respectively). These objects test robustness to previously unencountered compliances and edge sharpness in addition to novel edge curvature. 
    
\begin{figure}[t]
                \centering
                
                \begin{tabular}[b]{@{}cc@{}}
                        {\bf Plastic Banana} & {\bf Foam Brick} \\[0pt]
                        \includegraphics[width=.47\columnwidth]{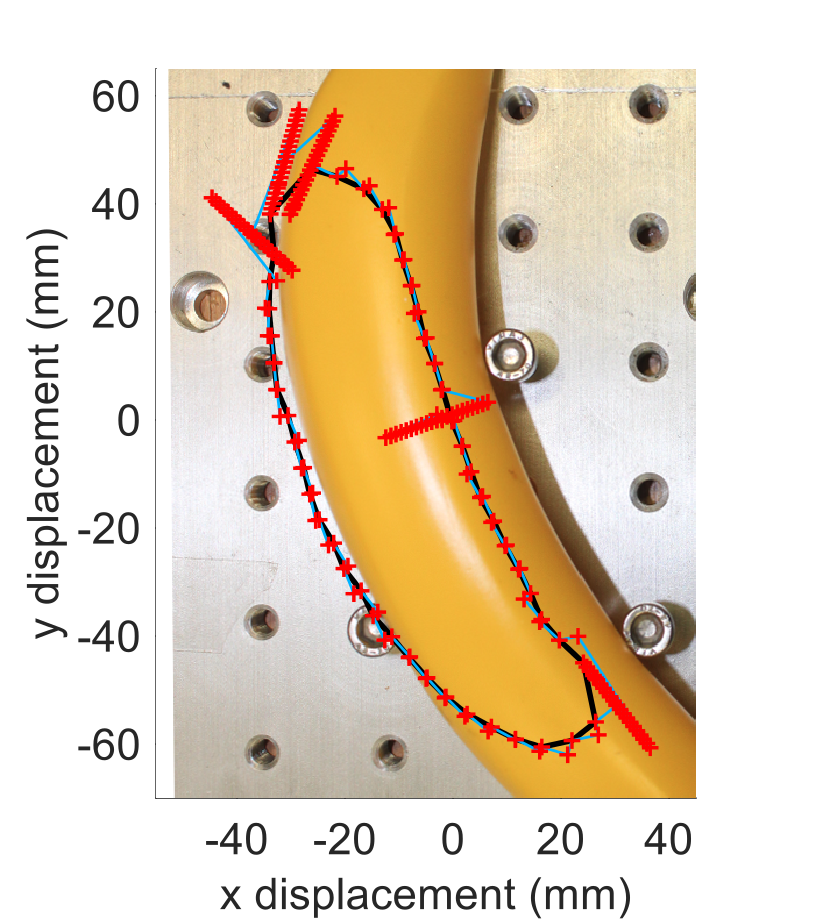} &
                        \includegraphics[width=.47\columnwidth]{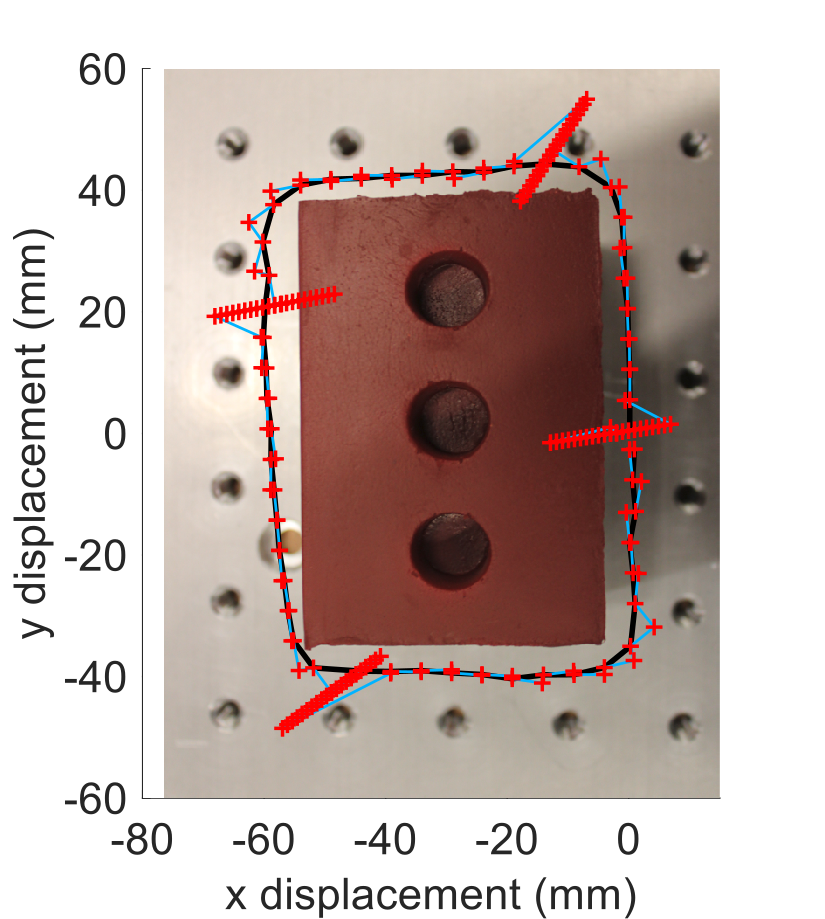}
                \end{tabular}
                  
                \caption{Contour following on a banana and foam brick. Red crosses are locations where the robot tapped, blue lines show the movement between taps, and the black line joins the locations identified as the object edge. Start location was (0,0) and movement is anticlockwise. (Note, location of background image is approximate.)}
                \label{unusual}
                \vspace{-15pt}
            \end{figure}

            The right-angles of the brick theoretically pose a challenge for this system due to the Exploration and Localisation lines being perpendicular; if the edge is being tracked perfectly the Exploration step will reach the corner and overshoot, tapping over free-space and attempting to find the minimum dissimilarity of the taps will result in a random location being chosen as the edge, leading the robot to add bad data to the model and to move in a random direction. The reason this is not reflected in the results (see \autoref{unusual}) is because the TacTip sensor is large and compliant, meaning that with this step size (10mm) the corners are effectively smoothed off (rather than moving starkly from on edge to off edge).
            
            The brick also adds the novelty of being made of very compliant foam which compresses when touched. This poses a challenge as the sensor readings vary with the compliance of the object.  This system is not specifically designed to handle differences in compliance, but owing to the use of $\phi$ as a representation of all variation not caused by displacement, this system is shown to not be affected by the compliance of the brick. 
            
            The plastic banana poses a different challenge, testing the resilience to changing edge sharpness. The sensor is started on a prominent edge of the banana, but this only runs along one side of the banana and the height of it changes as this side of the banana curves downwards at both ends. This means when following a 2D contour, maintaining the same height above the object as in this case, the edge will change from a sharp edge (along the length of the banana) into a very smooth edge (crossing the width of the banana) which is less an edge and more of a slope. This changes the sensor readings significantly.
            
            As can be seen in \autoref{unusual}, both the brick and banana were successfully traced. As expected, more data was needed than in the experiments with the circle and flower, but still only a maximum of 106 taps (5 lines) were used for the model.

%%%%%%%%%%%%%%%%%%%%%%%%%%%%%%%%%%%%%%%%%%%%%%%%%%%%%%%%%%%%%%%%%%%%%%%%%%%%%%%%
\section{Discussion \& Conclusion}
    This work has shown that online learning with a latent variable model and intelligent data collection policy is a data efficient and robust approach to tactile exploration with a TacTip tactile sensor.
    
    This method shows significant improvements in either accuracy or data efficiency over the previous methods with the TacTip. The histogram methods in ~\cite{Lepora2017} appear to use 180 taps for training, but are not very accurate in the tracking of the edge of objects. In our methods the highest number of taps in the model was 106, which is 40\% less data but with more accurate tracking of the object edges. The deep CNN in~\cite{Lepora2018} shows similar accuracy at tracking the edge (within $1\,\text{mm}$ of the edge) but used 2000 taps for training, which is 285 times more data than our most efficient case, and 19 times our least data efficient case. In addition to better data efficiency, the online method is able to cope with variations in the contour: the histogram methods do not generalise well, and while the deep CNN generalises well from its training data (a right angled edge) to everyday objects such as a banana with rounded sides, it was still not able to cope when there is no edge to follow at all (e.g. over the flat top of the banana), whereas the online method presented here is able to relearn the model to fully complete the loop.
    
    This work has shown that even with a simple architecture there are significant advantages in terms of data efficiency and robustness by using latent variable models and online learning methods in preference to offline learning methods when applied to tactile sensing tasks. Areas for future improvement on the contour-following task considered here include the use of arc localisation curves instead of straight lines to improve robustness to tighter curvatures, the use of more sophisticated exploration algorithms to improve the initial predictions at each step, and the use of raw camera pixels instead of taxel locations to retain more information about each tap and reduce the reliance on taxel detection algorithms.
    
    There are other opportunities in extending these methods to 3D contour or surface exploration where the collection of training data for offline learning methods is more challenging but the tasks potentially much closer to human capabilities to interact with the environment via touch. Similarly, the application to sliding motion instead of tapping will open up many possibilities, but is also a challenge to gather training data and construct a robust model, and it would be interesting to compare the performance with deep CNNs that are able to successfully generalise over sliding motion. In principle, there is no reason why the methods proposed here should not apply also to multi-fingered tactile robotic hands, opening up the possibility of online learning for the manipulation and exploration of complex objects.
    
    In conclusion, this work shows clear benefits to online learning with tactile sensors and lays the foundations for further research into its more widespread use. 




\section*{ACKNOWLEDGMENT}
% {\em Acknowledgements:} 
We thank Kirsty Aquilina and Noor Alakhawand for all their help.

% The preferred spelling of the word ÒacknowledgmentÓ in America is without an ÒeÓ after the ÒgÓ. Avoid the stilted expression, ÒOne of us (R. B. G.) thanks . . .Ó  Instead, try ÒR. B. G. thanksÓ. Put sponsor acknowledgments in the unnumbered footnote on the first page.

%%%%%%%%%%%%%%%%%%%%%%%%%%%%%%%%%%%%%%%%%%%%%%%%%%%%%%%%%%%%%%%%%%%%%%%%%%%%%%%%

% References are important to the reader; therefore, each citation must be complete and correct. If at all possible, references should be commonly available publications.

% \bibliographystyle{ieeetr}
% \bibliography{references} 
\bibliographystyle{IEEEtran}
\bibliography{IEEEabrv,main}

\end{document}